\documentclass{article}
\usepackage{tencent_ailab_tech_report}
\usepackage[colorlinks = true,
            linkcolor = blue,
            urlcolor  = blue,
            citecolor = blue,
            anchorcolor = blue]{hyperref}   
\usepackage{microtype}
\usepackage{hyperref}
\usepackage{xurl}
\usepackage{booktabs}
\usepackage{enumitem}
\usepackage{multicol}
\usepackage{CJKutf8}
\usepackage{amsmath}
\usepackage{siunitx}
\usepackage{floatflt}
\usepackage{graphicx}
\usepackage{booktabs}
\usepackage{wrapfig}
\usepackage{authblk}
\usepackage{lipsum}

\usepackage[ruled,vlined]{algorithm2e}  
\usepackage{mathrsfs,algorithmic}

\usepackage{microtype}
\usepackage{subfig}
\usepackage{multirow}
\usepackage{booktabs}   
\usepackage{pifont}     
\usepackage{hyperref}
\usepackage{amssymb} 

\usepackage{tabularx} 
\usepackage{booktabs}       
\usepackage{nicefrac}       
\usepackage{microtype}      
\usepackage{xcolor}         
\usepackage{bbm, dsfont}
\usepackage{array}
\usepackage{amsthm}
\usepackage{bm}
\usepackage{stfloats} 
\usepackage{mathtools}

\usepackage{array}
\usepackage{amsmath}
\usepackage{amssymb}
\usepackage{mathtools}
\usepackage{amsthm}

\usepackage[capitalize,noabbrev]{cleveref}
\usepackage{adjustbox} 

\theoremstyle{plain}

\theoremstyle{definition}

\theoremstyle{remark}

\usepackage{xcolor}

\colmfinalcopy

\usepackage{soul}
\usepackage{fontawesome5}
\usepackage{wrapfig}
\usepackage{caption}
\usepackage{colortbl}
\usepackage{mathabx}
\usepackage{arydshln}

\definecolor{nred}{RGB}{196, 38, 11}
\definecolor{ngreen}{RGB}{18, 141, 21}
\definecolor{nblue}{RGB}{41, 52, 190}
\definecolor{hzw}{RGB}{223, 97, 76}
\definecolor{lt}{RGB}{54, 89, 170}
\definecolor{tblue}{rgb}{0.867, 0.922, 0.969}
\definecolor{zlblue}{RGB}{196, 223, 251}

\newcommand{\ignore}[1]{}

\hypersetup{
    colorlinks=true,   
    linkcolor=purple,  
    citecolor=nblue,   
    filecolor=magenta, 
    urlcolor=nblue     
}

\crefname{section}{\S}{\S\S}
\Crefname{section}{\S}{\S\S}
\crefname{appendix}{appendix}{appendix}

\title{$\\ \\${\em FastMix:} Fast Data Mixture Optimization via Gradient Descent}

\author{%
Haoru Tan$^{1,2}\thanks{Personal Email: \textit{hrtan@eee.hku.hk} or \textit{tanhr2014@163.com}}$
~~~~~~~~~~~~~Sitong Wu$^{3}$
~~~~~~~~~~~~~Yanfeng Chen$^{2,\dagger}$
~~~~~~~~~~~~~Jun Xia$^{2}$
~~~~~~~~~~~~~Ruobing Xie$^2$
~~~~~~~~~~~~~Bin Xia$^{3}$
~~~~~~~~~~~~~~Xingwu Sun$^2$
~~~~~~~~~~~~~~Xiaojuan Qi$^{1,\dagger}$ \\
\vspace{10pt}
$^1$University of Hong Kong  \ \ \ \ \ \ $^2$Hunyuan LLM, Tencent \ \ \ \ \ \ $^3$Chinese University of Hong Kong\ \ \\ 
}

\colmfinalcopy 

\begin{document}

\maketitle

\begin{abstract} 
While large and diverse datasets have driven recent advances in large models, identifying the optimal data mixture for pre-training and post-training remains a significant open problem. 
We address this challenge with \textbf{\textit{\textsc{FastMix}}}, a novel framework that automates data mixture discovery while training only a \emph{single proxy model}. Instead of relying on predefined heuristics or resource-intensive simulations, \textsc{FastMix} jointly optimizes mixture coefficients and model parameters, substantially improving efficiency and scalability over prior approaches. At the core of \textsc{FastMix} is a reformulation of mixture selection as a \emph{bilevel optimization} problem. 
Under this reformulation, we show that optimizing mixture ratios is mathematically equivalent to assigning per-source loss weights under uniform source sampling. This embeds the mixture coefficients directly into the differentiable iterative optimization objective, enabling efficient, gradient-based optimization of both mixture and model. 
To solve the optimization problem, \textsc{FastMix} implements an approximate iterative optimization procedure, alternating between (i) updating model parameters on data sampled according to current mixture ratios (inner loop) and (ii) updating mixture ratios based on validation feedback (outer loop). 
Across pre- and post-training, \textsc{FastMix} outperforms baselines while drastically reducing search cost. Code (https://github.com/hrtan/fastmix)

\end{abstract}

\begin{figure*}[thtp]
\centering
\includegraphics[width=0.9088999094999\linewidth]{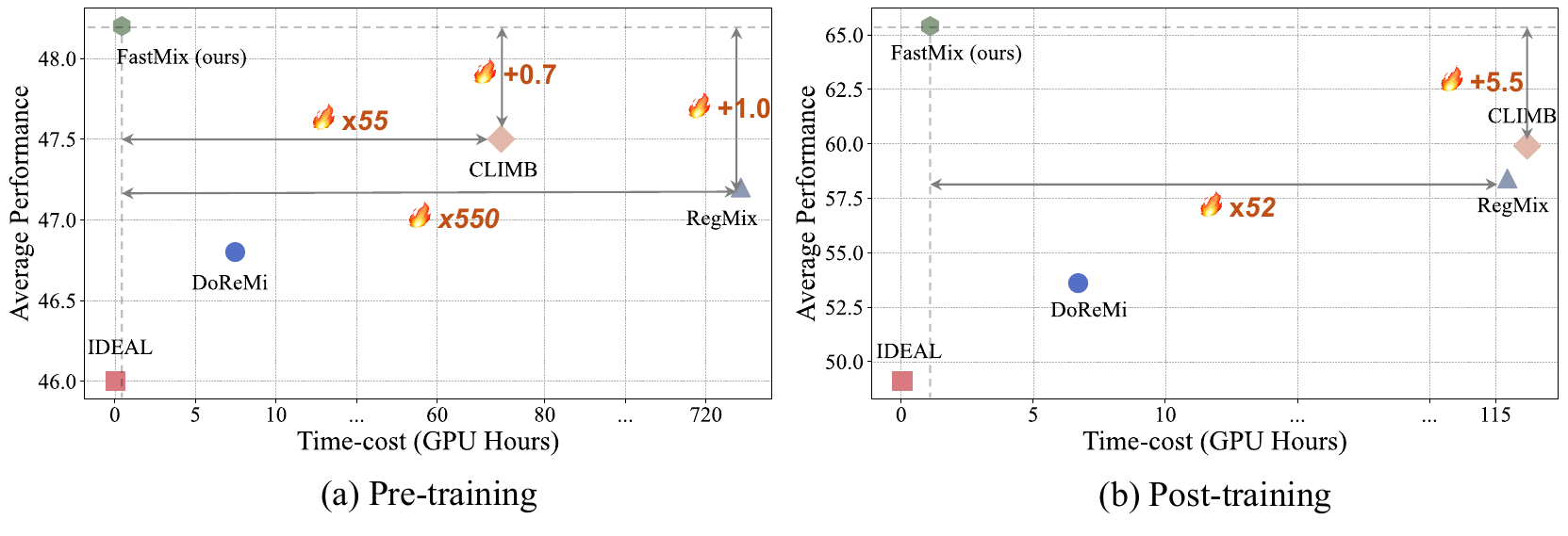}
\vspace{-0.3cm}
\caption{
        Average Performance versus Time-cost (GPU Hours) comparison for various data mixture strategies.
        \textbf{(a) Pre-training:} Our proposed \textit{\textsc{FastMix} (ours)} method achieves the highest performance with the lowest time-cost. The annotations highlight that it is up to \textit{55$\times$} more time-efficient than \textit{CLIMB} \citep{diao2025climb} and \textbf{550$\times$} more time-efficient than \textit{RegMix} \citep{liu2024regmix}, while providing a significant performance gain.
        \textbf{(b) Post-training:} In this setting, \textit{\textsc{FastMix} (ours)} again demonstrates state-of-the-art performance and time-efficiency, outperforming \textit{RegMix} with a \textbf{52$\times$} reduction in time-cost and gaining an additional \textbf{5.5} performance points over \textit{CLIMB}. This illustrates the superior trade-off between performance and time cost achieved by our method.
}
\label{fig:intro} 
\end{figure*}

\section{Introduction}

The performance of large-scale models \citep{yang2024qwen2,dubey2024llama,touvron2023llama,hu2024minicpm} depends critically on the data used for training. While large and diverse datasets have driven recent advances, identifying the optimal data mixture for pre-training \citep{shukor2025scaling} and post-training \citep{dong2023sftmixabilities} remains a significant challenge. 

Popular methods such as manual trial-and-error \citep{yang2023baichuan,tong2024cambrian} or proxy-based methods \citep{liu2024regmix,diao2025climb} often do not scale well as models grow larger. For example, proxy-based search methods such as RegMix \citep{liu2024regmix} and CLIMB \citep{diao2025climb} have demonstrated strong generalization and stability, yet they require training a large number of proxy models during the search. This results in prohibitive computational overhead, making mixture optimization increasingly impractical as both models and datasets continue to expand. The central question is thus: how can we efficiently determine effective data mixtures for large-scale training?

We address this challenge with \textsc{\textsc{FastMix}}, a novel framework that automates data mixture discovery while training only a \emph{single proxy model}. Instead of relying on predefined heuristics or resource-intensive simulations, \textsc{FastMix} jointly optimizes mixture coefficients and model parameters, substantially improving efficiency and scalability over prior approaches. 
At the core of \textsc{FastMix} is a reformulation of mixture selection as a weighted \emph{bilevel optimization} problem in Eq.\eqref{eq: our problem}. Specifically, we show that optimizing mixture ratios is mathematically equivalent to assigning per-source loss weights under uniform source sampling. This reparameterization embeds the mixture coefficients directly into the differentiable iterative optimization objective, enabling efficient, gradient-based optimization of both mixture and model. 
To solve the optimization problem \citep{maclaurin2015gradient,franceschi2018bilevel}, \textsc{FastMix} implements an approximate iterative optimization procedure, alternating between (i) updating model parameters on data sampled according to current mixture ratios (inner loop) and (ii) updating mixture ratios based on validation feedback (outer loop) via a gradient-based optimizer \citep{adam}.

Extensive evaluations demonstrate that \textsc{FastMix} optimizes data mixtures across model scales and tasks in both pre-training and post-training, outperforming baselines at a fraction of the computational cost (See Fig. \ref{fig:intro}).  In pre-training, it delivers a top average score of {48.2} and rank {1} across {14} benchmarks (best on {9}) with just \textbf{1.3} GPU-hours, achieving \(\times\)\textbf{550} faster than RegMix \citep{liu2024regmix} and \(\times\)\textbf{55} than CLIMB \citep{diao2025climb}. In post-training (SFT), a math-tuned mixture generalizes to coding and STEM-QA, reaching {65.4} (\(\textbf{+5.5}\) over next best) in \textbf{2.2} GPU-hours versus more than {\(\textbf{115}\)} GPU-hours for CLIMB/RegMix. Overall, \textsc{{FastMix}} makes mixture optimization practical and scalable for next-generation large models.

\section{Related Work}

The rapid progress of large models \citep{dubey2024llama,touvron2023llama,allal2024SmolLM,yang2023baichuan,yang2024qwen2technicalreport} relies heavily on strategically mixing data from diverse sources, spanning languages \citep{yang2023baichuan}, modalities \citep{gunasekar2023textbooks,yang2024qwen2}, and difficulty levels \citep{he2025deepmath}. This \emph{data mixture problem} \citep{ge2024data} presents fundamental challenges not only in pre-training \citep{shukor2025scaling,dubey2024llama,yang2024qwen2} but also in post-training \citep{dong2023sftmixabilities,ming2025ideal,tong2024cambrian}. Early practice largely relied on manual heuristics, which lack standardization and often fail to generalize across settings. More recently, optimization-based approaches \citep{xie2024doremi,fan2023doge,liu2024regmix} have been introduced to automate mixture selection.

\textbf{Proxy-based methods} \citep{xie2024doremi,liu2024regmix,diao2025climb} adopt a two-phase design in which a proxy model is trained under candidate mixtures and its performance is used to infer optimal sampling ratios. For example, DoReMi \citep{xie2024doremi} trains a small proxy to adjust domain weights based on relative losses, then reuses the optimized ratios to train a larger model. RegMix \citep{liu2024regmix} scales this idea by training hundreds of proxy models under different ratios, fitting a regression model on the resulting mixture-performance pairs, and extrapolating the optimal mixture. CLIMB \citep{diao2025climb} improves efficiency by iteratively refining the search region, reducing the number of proxy models required. Other works \citep{ye2024data,shukor2025scaling,kang2024autoscale} study cross-scale transfer: \citet{shukor2025scaling} provide theoretical and empirical evidence that mixtures found on small models generalize to larger ones, while \citet{ye2024data,kang2024autoscale} report functional relationships between mixture proportions and performance.

In contrast, \textbf{dynamic methods} \citep{chen2024aioli,ming2025ideal,albalak2023odm} remove the separate search phase by adjusting mixtures on the fly. IDEAL \citep{ming2025ideal}, for instance, leverages influence functions \citep{koh2017understanding} to estimate domain contributions to downstream performance and to dynamically rebalance training data.

Overall, proxy-based methods such as RegMix and CLIMB generally achieve stronger and more stable performance than dynamic approaches, but at substantial computational cost. Our method, \textsc{FastMix}, preserves the reliability of proxy-based optimization while cutting search time from hundreds of GPU-hours to nearly one, achieving both higher efficiency and stronger generalization.

\section{{FastMix}}

\subsection{Problem reformulation with reparameterization}

\paragraph{Data Mixture as a Bi-level Optimization Problem.} 
Formally, data mixture optimization can be posed as a bilevel optimization problem.
Let \(D=\{D_1,\dots, D_k\}\) be a collection of data sources (or clusters),
and let \(\alpha\in A\subset\mathbb{R}^k\) denote the mixture weights,
where the feasible set \(A\) is the probability simplex
(\(\alpha_i\ge 0\) and \(\sum_{i=1}^k \alpha_i=1\)).
Given mixture \(\alpha\) and model parameters \(w\), 
the training objective is \(\mathcal{L}_{\text{train}}(D, w \mid \alpha)\). 
Let \(w^*(\alpha)\) be the parameters obtained by (approximately)
optimizing this training objective under \(\alpha\).
The target is to find mixture weights \(\alpha^*\) that minimize the validation loss, i.e., 
\(\mathcal{L}_\text{target}(w) = \ell_{\text{val}}(V, w)\) evaluated at \(w^*(\alpha)\):  
\begin{equation}  
\min_{\alpha} \,\, \mathcal{L}_\text{target}\Big(w^*(\alpha)\Big) ~~~~~\text{s.t.} ~~~~~w^*(\alpha) = \arg\min_{w} \mathcal{L}_\text{train}\Big(D, w |\alpha\Big), ~~~ \sum_{i=1}^{k} \alpha_i = 1, ~~~\alpha_i \geq 0. 
\end{equation} 
where the inner-loop aims to find the optimal model weights $w^*(\alpha)$ by minimizing the training loss on the dataset given mixture weights $\alpha$. The outer-loop then seeks to optimize these mixture weights $\alpha$ to minimize the model's final loss on target tasks.

While the bi-level formulation is conceptually appealing, it is difficult to solve in practice.
The crux is handling the mixture weights \(\alpha\).
Unlike model parameters \(w\), which admit efficient gradient-based updates,
mixture (sampling) ratios are typically non-differentiable, precluding end-to-end backpropagation.
Consequently, practitioners resort to greedy heuristics or policy-gradient (score-function) updates to adjust \(\alpha\).
These procedures are sample-inefficient and scale poorly with the number of data sources, turning mixture search into a dominant computational bottleneck.

\paragraph{Differentiable Formulation.}

Through a simple reparameterization, we recast the original bilevel problem into a mathematically equivalent, fully differentiable objective. The key idea is to replace stochastic sampling by mixture ratios with per-source, differentiable loss weights applied under uniform sampling, so that each source’s contribution is controlled continuously via its weight, yielding the following formulation: 
\begin{equation}
\min_{\alpha} \,\, \mathcal{L}_\text{target}\Big(w^*(\alpha)\Big)
~~~~~\text{s.t.} ~~~w^*(\alpha) = \arg\min_{w} \sum_{i=1}^k  \alpha_i \mathcal{L}_\text{train}\Big(D_i, w\Big), ~~~ \sum_{i=1}^{k} \alpha_i = 1, ~~~\alpha_i \geq 0, 
\label{eq: our problem}
\end{equation} 
where \(\mathcal{L}_{\text{train}}(D_i, w)\) denotes the model’s training loss on source \(D_i\), computed under \emph{uniform source sampling} (each source selected with probability \(1/k\)). 
The inner-loop finds the optimal model weights, $w^*(\alpha)$, by minimizing a weighted sum of the training losses from $k$ different data domains. The data mixture weight $\alpha_i$ serves as the weight for each domain's loss. The outer-loop then aims to optimize these proportions $\alpha$ to minimize the model's loss on target tasks. 
This reparameterization is key: rather than treating mixture ratios as non-differentiable sampling probabilities, we reinterpret them as continuous coefficients that scale each source’s loss.
Consequently, the mixture weights \(\boldsymbol{\alpha}=(\alpha_1,\ldots,\alpha_k)\) are fully differentiable and amenable to gradient-based optimization.
Standard optimizers (e.g., SGD or Adam) can then jointly update the model parameters and the data weights, enabling efficient end-to-end training.

\textit{Proof of equivalence.} 
Let \(D=\bigcup_{i=1}^k D_i\) denote the union of \(k\) data sources (or clusters),
and let \(\alpha=(\alpha_1,\dots,\alpha_k)\) be mixture weights with \(\sum_i \alpha_i=1\), \(\alpha_i\ge0\).
To sample a training example \(x\), first draw a source index \(i \sim \mathrm{Cat}(\alpha)\), then sample \(x \sim D_i\).
The training loss under this mixture sampling is
\begin{equation} 
\mathcal{L}_{\text{train}}(D,w \mid \alpha)
= \mathbb{E}_{i \sim \mathrm{Cat}(\alpha)} \, \mathbb{E}_{x \sim D_i}\big[\ell(x,w)\big]
= \sum_{i=1}^k \alpha_i \, \mathcal{L}_{\text{train}}(D_i,w),
\end{equation}
where \(\ell(x,w)\) is the per-example loss and
\(\mathcal{L}_{\text{train}}(D_i,w)=\mathbb{E}_{x \sim D_i}[\ell(x,w)]\) is the expected loss on source \(D_i\).
Thus, under mixture sampling, the expected training loss is a convex combination of the per-source losses, with coefficients given by the mixture ratios.

\subsection{How to obtain better generalization performance?}

{Like most AutoML algorithms, \textsc{FastMix}  requires a search target, typically defined as a performance metric on a held-out validation set. However, relying on validation performance alone can lead to overfitting to quirks of the validation data and limited transferability to new scenarios. To improve generalization, we propose two complementary strategies: (i) entropy-based regularization to encourage diversity among mixture weights, and (ii) incorporating training loss into the search target to balance validation and training signals.}

\textbf{Entropy-based regularization.}  
{Entropy regularization prevents the mixture distribution from collapsing onto a narrow subset of data sources. Given mixture weights $(\alpha_1, \dots, \alpha_k)$ across $k$ sources, we add the penalty 
$\mathcal{R}_{\text{entropy}} = \sum_{i=1}^{k} \alpha_i \log \alpha_i$. 
Minimizing this term discourages overly peaked distributions, promoting more uniform weight allocation. This reduces sensitivity to spurious validation patterns and improves robustness by leveraging multiple data sources.}

\textbf{Training loss as an auxiliary target.}  
{We further integrate the training loss into the search objective to complement the validation signal. While the validation term reflects out-of-sample generalization, the training term measures how effectively the model fits the mixture as a whole. Combining the two reduces over-reliance on the limited validation set and guides the search toward mixture ratios that generalize more reliably across both in-domain and out-of-domain data.}

\textbf{Joint objective.}  
{Together, entropy regularization and the auxiliary training loss yield the following search objective:}
\begin{equation}  
\vspace{-0.2cm}
    \mathcal{L}_{\text{target}}(w) = \ell_{\text{val}}(w) ~+~ \beta \, \mathcal{L}_{\text{train}}(w) ~+~ \lambda \sum_{i=1}^{k} \alpha_i \log \alpha_i ,
\end{equation} 
{where $\beta \geq 0$ and $\lambda \geq 0$ are trade-off hyperparameters. Empirically, $\lambda$ is set to a small value (e.g., $10^{-5}$) to encourage diversity without dominating the optimization, while $\beta$ is most effective at moderate values (e.g., $0.1$). We provide a detailed sensitivity analysis of these hyperparameters in our ablation studies. 
Overall, these two strategies substantially improve the generalization ability of \textsc{FastMix}, enabling it to discover mixtures that not only perform strongly on validation benchmarks but also transfer robustly to broader real-world applications. }

\subsection{Optimization} 

{Although the reparameterized formulation enables end-to-end differentiation over both model parameters and data mixtures, the resulting bilevel problem is still difficult to solve directly. Accordingly, we adopt an iterative procedure (Alg.~\ref{alg}) that alternates between updating the model parameters and the mixture weights \citep{maclaurin2015gradient,liu2018darts,pedregosa2016hyperparameter,franceschi2018bilevel}. The two key steps are outlined below. }

\textbf{(i) Inner loop (network parameter update).}  
{Given current mixture weights $\alpha^t$, the model parameters $w$ are updated for $n_1$ steps via stochastic gradient descent (SGD) to minimize the weighted training loss $\mathcal{L}_\text{train}$ :}
\begin{equation}  
w^{t+1} \gets w^{t} - \eta_w^t \frac{\partial \Big( \sum_{i=1}^k  \alpha^t_i \mathcal{L}_\text{train}(D_i, w^t) \Big)}{\partial w^t},
\end{equation}
{where \(\mathcal{L}_{\text{train}}(D_i, w)\) denotes the model’s training loss on source \(D_i\), computed under \emph{uniform source sampling} (each source selected with probability \(1/k\)). This is repeated for $n_1$ iterations. Other gradient-based optimizers, such as Adam \citep{adam}, are compatible with our framework. After $n_1$ updates, we denote the resulting parameters as $w^{t+n_1}$. } 

\textbf{(ii) Outer loop (mixture weight update).}  
{The mixture weights $\alpha^t$ are then updated using validation feedback $\mathcal{L}_\text{target}$. Specifically, the model is trained for $n_2$ iterations with the previous mixture weights $\alpha^t$, and the resulting parameters $w^{t+n_2}$ are evaluated on the validation loss $\mathcal{L}_\text{target}$. The mixture weights are updated as:}
\begin{equation}  
\alpha^{t+1} \gets \alpha^t - \eta_\alpha^t  \frac{\partial \mathcal{L}_\text{target}\big(w^{t+n_2}\big)}{\partial \alpha^t}, 
\end{equation}  

{In effect, $\alpha^{t+1}$ is updated according to how the validation loss responds after $n_2$ steps of training under $\alpha^t$. This naturally assigns larger weights to data sources that contribute more to improving validation performance. A key consideration is how the gradient is estimated, since this directly impacts both the direction of updates and the efficiency of the search.}  

{In the special case $n_2 = 1$ with SGD updates, the gradient of the validation loss with respect to $\alpha^t_i$ yields a closed-form solution:}
\begin{align}
\frac{\partial \mathcal{L}_\text{target}\big(w^{t+1}\big)}{\partial \alpha^t}
= \frac{\partial \mathcal{L}_\text{target}(w^{t+1})}{\partial w^{t+1}} 
\cdot \frac{\partial w^{t+1}}{\partial \alpha^t_i} 
= - \eta_w^t \, \nabla_w \ell_\text{val}(V, w^{t+1}) 
\cdot \nabla_w \mathcal{L}_\text{train}(D_i, w^t),
\label{eq:update}
\end{align}
{where $D_i$ denotes the $i$-th training source. This shows that per-source training losses directly shape the mixture gradients. 
The following derivation shows why the formula holds. Under the SGD update rule, the weights $w$ at time $t+1$ are updated based on the gradient of the loss function with respect to the mixture coefficients $\alpha_i^t$: $w^{t+1} = w^t - \eta^t_w \nabla_{w} [ \sum_{i=1}^k \alpha^t_i \, \mathcal{L}_{\text{train}}(D_i,w^t) ]$. Taking the derivative of $w^{t+1}$ with respect to $\alpha^t_i$, we get:
$\frac{\partial w^{t+1}}{\partial \alpha^t_i} = \frac{\partial}{\partial \alpha^t_i} \left[ w^t - \eta^t_w \nabla_w \left( \sum_{j=1}^k \alpha^t_j \, \mathcal{L}_{\text{train}}(D_j,w^t) \right) \right]$. 
Since $w^t$ is independent of $\alpha^t_i$, the derivative of the first term is zero. Due to the linearity of the derivative and the sum, only the term corresponding to $\alpha^t_i$ remains, hence,  $\frac{\partial w^{t+1}}{\partial \alpha^t_i} = - \eta^t_w \nabla_{w} \mathcal{L}_{\text{train}}(D_i,w^t)$. }

{The formulation in Eq.\eqref{eq:update} can be intuitively understood as follows: 
The gradient with respect to \(\alpha_i\) is proportional to the \emph{alignment} between
(i) the validation gradient \(\nabla_w \ell_{\text{val}}(V,w^{t+1})\) and
(ii) the training gradient from source \(D_i\), \(\nabla_w \mathcal{L}_{\text{train}}(D_i,w^{t})\).
If these gradients are aligned (positive dot product), the derivative
\(-\,\eta_w^t\,\nabla_w \ell_{\text{val}}\!\cdot\!\nabla_w \mathcal{L}_{\text{train}}(D_i,w^{t})\) is negative, so a gradient-descent step on \(\alpha_i\) \emph{increases} its weight, emphasizing sources whose updates also reduce the validation loss.
If they are opposed (negative dot product), the derivative is positive and a step \emph{decreases} \(\alpha_i\), down-weighting sources that harm validation performance.
Near-orthogonality yields small updates.
Thus, the procedure reallocates mass toward data sources whose training signals most effectively improve the validation objective.}

{When $n_2 > 1$, deriving a closed-form gradient becomes intractable, requiring finite-difference approximations or similar techniques, which are often unstable and inefficient. In contrast, $n_2 = 1$ admits a closed-form gradient that is both computationally efficient and empirically effective.

\begin{algorithm}[tp]
\caption{\textsc{FastMix} Optimization Algorithm}
\label{alg}
\begin{algorithmic}[1]
\STATE   \textbf{Initialize} model parameters $w^0$, mixture weights $\alpha^0$, inner-loop duration $n_1$ and outer-loop duration $n_2$. 
\FOR{$t = 0, 1, \dots, T-1$}  
    \IF{$(t) \bmod n_1 \neq 0$}
        \STATE   \textcolor{blue}{// Inner loop: update model parameters (\textit{e.g.}, via the SGD optimizer, and we can change this update rule to other optimizers, like Adam \citep{adam})}
        \STATE $w^{t+1} \gets w^{t} - \eta_w^t \frac{\partial [ \sum_{i=1}^k  \alpha^t_i \mathcal{L}_\text{train}(D_i, w^t) ]}{\partial w^t},$
    \ELSE
        \STATE  \textcolor{blue}{// Outer loop: update mixture weights (\textit{e.g.}, via the SGD optimizer, and we can change this update rule to other optimizers, like Adam \citep{adam})}
        \STATE $\alpha^{t+1} \gets \alpha^t - \eta_\alpha^t  \frac{\partial \mathcal{L}_\text{target}\big(w^{t+n_2}\big)}{\partial \alpha^t}$
    \ENDIF
\ENDFOR 
\STATE \textbf{Output}: the optimized mixture weight $a^\text{final}$ after the final outer loop update. 
\end{algorithmic}
\end{algorithm}

\section{Experiments}

To comprehensively evaluate the effectiveness of our proposed framework, we conduct experiments on data mixture optimization across different stages of large language model (LLM) training, including both pre-training and post-training. The compared methods cover a wide spectrum of approaches, ranging from human expert tuning to proxy-based search methods such as DoReMi \citep{xie2024doremi}, RegMix \citep{liu2024regmix} and CLIMB \citep{diao2025climb}, and dynamic methods, including ODM \citep{albalak2023odm} and IDEAL \citep{ming2025ideal}. The subsequent sections are organized as follows: Section~\ref{sec:pretraining} presents results on pre-training mixture optimization. Section~\ref{sec:posttraining} reports experiments in post-training settings.

\subsection{Pre-training Stage Experiments}  
\label{sec:pretraining}

\textbf{Setups.} Following prior work~\citep{liu2024regmix}, we conduct our experiments on the Pile dataset~\citep{pile}, focusing on the 17 uncopyrighted subsets available on HuggingFace. 
For mixture optimization in the pre-training stage, we employ small proxy models (e.g., 1M parameters) trained on up to 1B tokens. 
To test the method's generalization ability, consistent with~\citet{liu2024regmix}, we use the loss on a representative and diverse part of the training data (the Pile-cc sub-set \citep{pile}) as the search target. 
For \textsc{FastMix}, we employ only a single proxy model, whereas RegMix uses 512 by following \citep{liu2024regmix} proxy models and CLIMB uses 64 \citep{diao2025climb}. For the Human Heuristic baseline, we directly adopt the manually tuned mixture configuration reported in \citep{liu2024regmix} to ensure fairness. 
After the search stage, we use the mixture configurations obtained by each method to train a 1B-parameter model on 25B tokens. 
For evaluation, we focus on the accuracy of the pretrained model on a suite of downstream task benchmarks, including Social IQA~\citep{sap2019socialiqa}, HellaSwag~\citep{zellers2019hellaswag}, PiQA~\citep{bisk2020piqa}, et.al. 
In addition, we also examine the time cost incurred by different methods during the search stage.

\begin{figure*}[tp]
\centering
\includegraphics[width=0.99988999094999\linewidth]{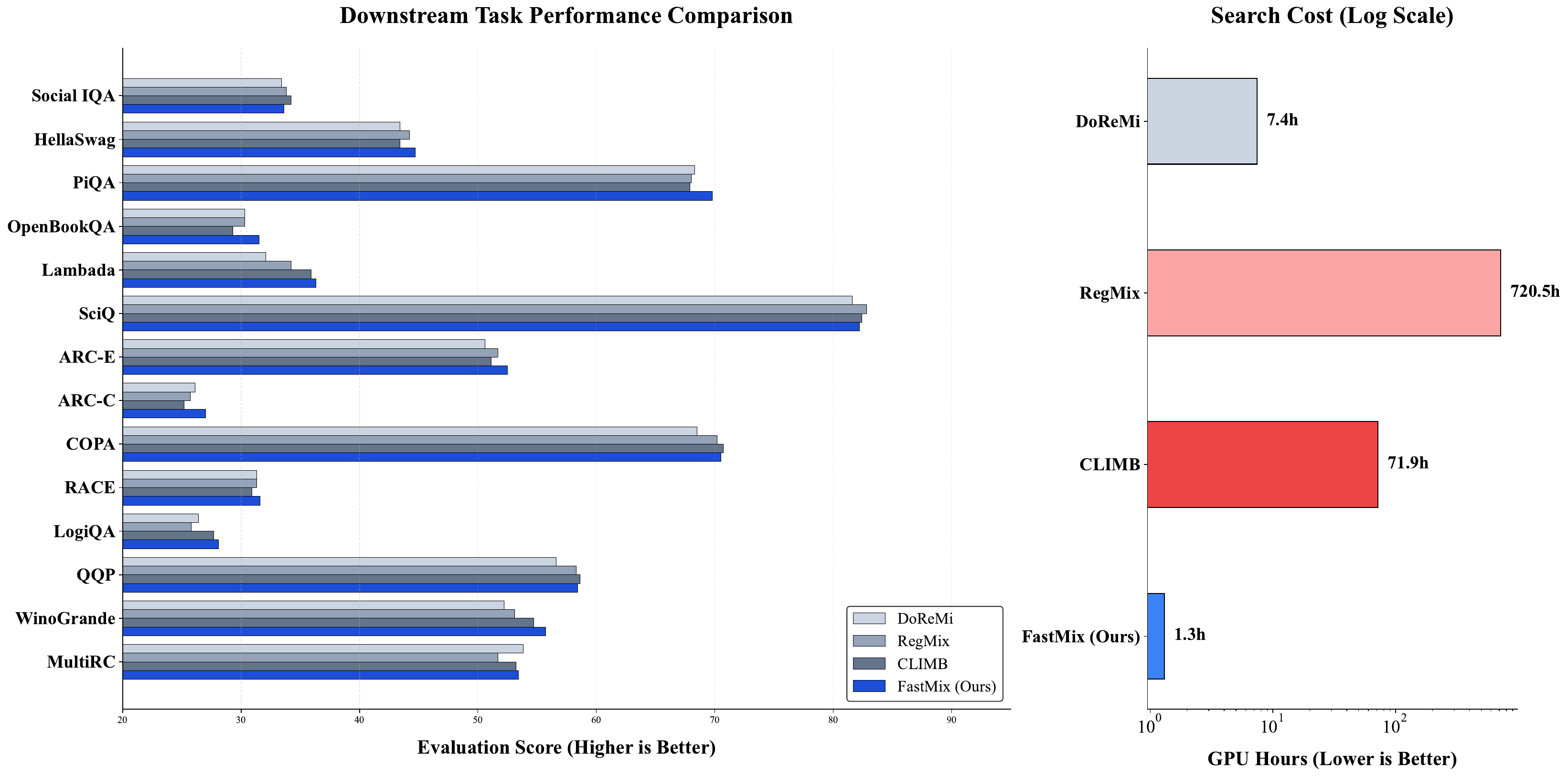}
\vspace{-0.3cm}
\caption{Comparative evaluation of different data mixture strategies in the context of large-scale pretraining, examining their impact on both downstream task performance and training efficiency. 
}
\label{fig:downstream_perf_our} 
\end{figure*}

\textbf{Results.} As shown in Figure~\ref{fig:downstream_perf_our}, our proposed method, \textsc{FastMix}, demonstrates significant advantages in both downstream task performance and computational efficiency compared to existing data mixture strategies. It achieves the highest average performance score of 48.2 and the best average rank of 1 across all 14 downstream benchmarks, outperforming strong baselines including CLIMB (47.5) and RegMix (47.2). This top ranking underscores its consistent and robust generalization capabilities, further evidenced by its leading results on 9 of the 14 individual tasks. Most notably, \textsc{FastMix} offers a dramatic improvement in search efficiency, requiring only 1.3 GPU-hours to identify the optimal mixture. This is orders of magnitude faster than other automated methods, such as CLIMB (71.9 GPU-hours) and RegMix (720.5 GPU-hours), validating the efficacy of our single proxy model and gradient-based optimization approach. Collectively, these results confirm that \textsc{FastMix} not only discovers superior data mixture configurations but also drastically reduces the computational overhead of the search process, offering a scalable and practical solution for large-scale model training.

\subsection{Post-training  Stage Experiments}
\label{sec:posttraining}

\textbf{Setups.} Building on our pre-training success, we next validated \textsc{FastMix} in the post-training stage, aiming to optimize data mixtures for specialized tasks on the Qwen2.5-Math-Instruct 7B model \citep{hui2024qwen25math}. For this study, we sourced supervised fine-tuning (SFT) data from eight distinct domains, including Math (OpenR1-Math-220k \citep{openr1_math_220k}), Code (the programming-related subset from the OpenThoughts-114K \citep{guha2025openthoughtsdatarecipesreasoning}), Dialogue (ShareGPT \citep{ryokoai_sharegpt52k_2023}), and STEM (Platypus \citep{lee2023platypus}). Our optimization  search objective  was a 1:1 weighted sum of scores from two mathematical benchmarks, the simpler GSM8K \citep{cobbe2021trainingverifierssolvemathgsm8k} and the more challenging  gaokao2023en \citep{gaokao2023_math_en}. To evaluate the model's generalization capabilities, we extended our test suite beyond math (MATH \citep{hendrycks2021measuringmathbench}, AIME-24 \citep{jia2024aime}) to include tasks in coding (LiveCodeBench-v2 \citep{jiang2024livecodebench}) and STEM question-answering (GPQA-Diamond \citep{rein2023gpqa}). 
A significant challenge in the post-training setting is the absence of very small (e.g., 10M parameter) proxy models. Therefore, we had to conduct our search using proxy models of approximately 1 billion parameters (Qwen2.5-1.5B-Instruct \citep{qwen2025qwen25technicalreport}), with evaluation performed on larger models (7B). This constraint exposed a critical limitation of resource-intensive methods \citep{liu2024regmix,diao2025climb}, which require training hundreds of proxy models. Given the immense computational cost, our cluster was unable to support hundreds of full 1B-model training runs, so we had to reduce the number of proxy models for both RegMix and CLIMB to just 64. In contrast, \textsc{FastMix}'s reliance on a single proxy model enabled it to operate efficiently within these resource limitations, highlighting its superior scalability for larger-scale tasks.

\textbf{Results.} In the post-training (SFT) stage, the advantages of \textsc{FastMix} are further solidified, demonstrating an even more dominant performance as shown in Figure \ref{fig:post}. Our method achieved the highest score across all four benchmarks spanning mathematics, coding, and general question-answering, resulting in a superior average performance of 65.4 and a top rank of 1, by a significant 5.5 point lead over the next best method, CLIMB (59.9) \citep{diao2025climb}. Crucially, these results highlight the exceptional generalization capability of \textsc{FastMix}. While all automated methods used performance on mathematics benchmarks (GSM8K and gaokao2023en) as the guidance signal for optimization, \textsc{FastMix} not only excelled in the math domain but also achieved the best performance on LiveCodeBench (coding) and GPQA-Diamond (STEM QA). This strongly indicates that the data mixture identified by \textsc{FastMix} avoids overfitting to the optimization signal and instead fosters a more fundamental and comprehensive improvement in the model's capabilities, all while maintaining remarkable efficiency by completing its search in just 2.2 GPU hours, substantially faster than RegMix (115.9 hours) and CLIMB (117.4 hours).

\begin{figure*}[tp]
\centering
\includegraphics[width=0.99988999094999\linewidth]{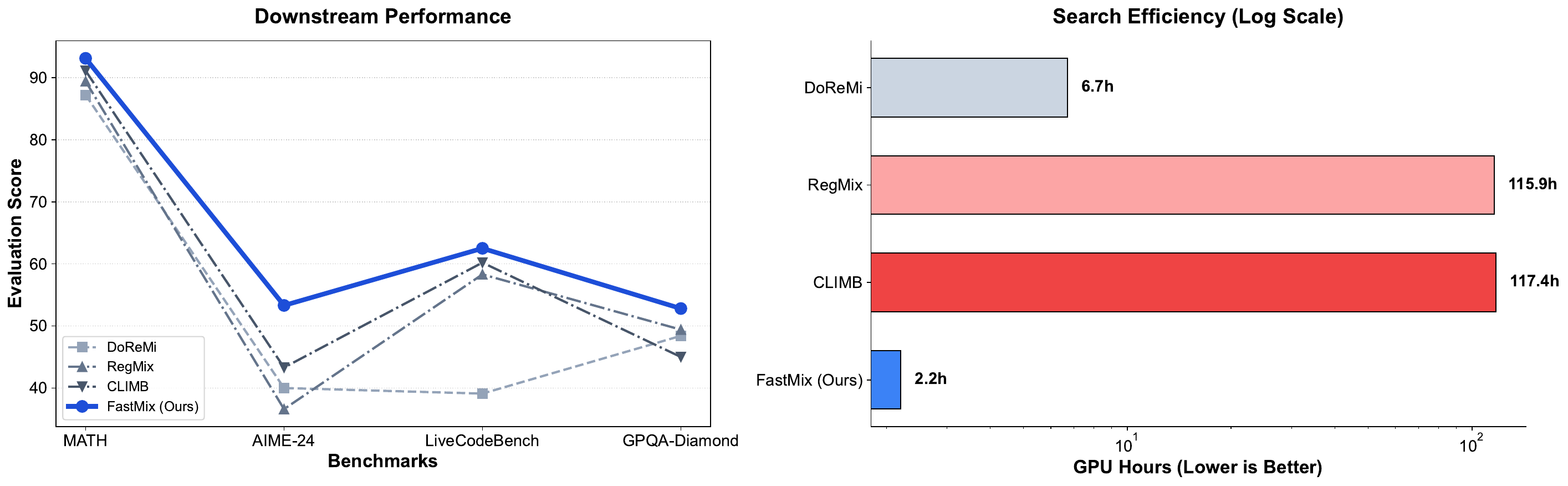}
\vspace{-0.3cm}
\caption{Comparative evaluation of different data mixture strategies in the context of large-scale post-training (SFT), examining the efficiency and downstream task performance. 
}
\label{fig:post} 
\end{figure*}

\subsection{Tip: The painful lesson of no free lunch}

In this sub-section, we conducted some very necessary discussions. Some of the conclusions are derived from the experience in the industrial development process and may be quite different from the simple and clean conclusions obtained from academic data sets.

\textbf{Non-differentiable targets.}  Our optimization algorithm is designed for settings where both $\mathcal{L}_\text{target}$ and $\mathcal{L}_\text{train}$ are differentiable. However, in practice, non-differentiable situations may arise. We discuss two representative cases below.  
One common challenge arises when the objective function is non-differentiable, such as when validation performance is measured by discrete metrics (e.g., accuracy) rather than a smooth loss. In such cases, we propose using a differentiable proxy objective, for instance, the supervised fine-tuning (SFT) loss for question-answering tasks, which provides a smooth surrogate while remaining aligned with the discrete evaluation metric. This approach has proven to be highly effective in practice.

\textbf{Black-box gradient estimators.}  We conducted extensive experiments, and the results indicate that it is highly challenging to estimate gradients for non-differentiable metrics using methods like finite differences or Simultaneous Perturbation Stochastic Approximation (SPSA). Convergence is rarely achieved, particularly on industrial datasets. 
We attribute this difficulty to two primary reasons. First, SPSA relies heavily on hyperparameter tuning for gradient estimation, and its estimation accuracy is inherently poor. Second, while finite differences depend on introducing small perturbations to the parameters, non-differentiable metrics often require substantial perturbations to show even marginal changes. This renders the gradient estimates extremely noisy. Furthermore, the finite difference method requires perturbing each source individually; this process is highly inefficient and fails to scale to a large number of sources. Consequently, we suggest exercising extreme caution when considering black-box metrics as optimization objectives for FastMix.

\textbf{Long outer-loop horizons.}  
Another challenge arises when the outer-loop duration parameter $n_2$ is greater than one. In this case, computing the gradient of the mixture weights becomes intractable. Without constraints on $n_2$, one would either need to rely on built-in mechanisms in PyTorch \citep{pytorch}, such as backpropagation-through-time (BPTT), which quickly becomes prohibitively memory-intensive in large-model settings, or fall back on general gradient-estimation techniques such as finite differences, which again are slow and unstable. To avoid these pitfalls, we restrict $n_2=1$ whenever possible, which not only yields a closed-form gradient but also delivers the most stable and efficient optimization behavior.

\textbf{About the regularization terms.} On simple and clean academic datasets, such a straightforward approach can be considered to prevent the optimization from collapsing onto just one or a few sources, which is a common issue in most current data-mixing algorithms. However, our extensive development experience with industrial data indicates that regularization terms may not be particularly effective. Instead, the most robust solution is to enforce strict oversampling ratio constraints across all sources (for instance, capping the up-sampling at three times the original size).

\textbf{About the small proxy model.} In industrial scenarios, caution should be exercised when relying on small surrogate models (smaller than 0.5B) to determine hyperparameters, such as data-mixing ratios. Based on our extensive experimentation with industrial data, small surrogate models exhibit significant limitations. First, they suffer from convergence instability, which often yields highly noisy mixing ratios; this issue appears inherently tied to model scale rather than the algorithm itself, as we observed the same phenomenon even when using RegMix as an oracle. Second, discrepancies in model capacity and architecture naturally lead to distinct biases toward different data sources.

\textbf{About the search target data.} In this study, we adhere to the experimental setup of RegMix, utilizing the loss on the Pile-cc validation set as our optimization target. In industrial development, however, practitioners typically maintain proprietary validation sets distinct from the test set. As suggested previously, open-ended questions within these sets can be formulated into SFT data to compute SFT loss. Crucially, we identify a major bottleneck in pre-training: pre-training sequences are typically long, whereas SFT data is significantly shorter. This structural discrepancy causes the gradients computed on these two data types to diverge drastically, ultimately leading to the failure of FastMix. To mitigate this issue, a straightforward yet highly effective solution is to concatenate multiple SFT sequences to align their lengths with the pre-training data.

\section{Conclusion}
We introduced \textsc{FastMix}, an efficient framework for discovering data mixtures for large-model training. Our key contribution is a weighted bilevel reformulation of mixture selection: via a reparameterization, optimizing sampling ratios becomes equivalent to learning per-source loss weights, enabling mixture coefficients to be differentiable. This permits joint, gradient-based optimization of both the model and the mixture using a single proxy model rather than hundreds. Across pre-training and post-training, FastMix delivers superior accuracy with orders-of-magnitude lower search cost, making data mixture optimization practical, scalable, and robust for next-generation LLMs.

\section{Future Works}
FastMix also exhibits certain limitations and areas for future exploration. First, its current one-step, short-horizon outer-loop update mechanism introduces a degree of greediness, making the algorithm somewhat sensitive to data noise. Second, we observed intriguing search dynamics during the optimization process: many data sources exhibit a competitive, time-evolving relationship. Certain sources prove vital in the early stages, whereas the most critical sources dominate only after prolonged training. This phenomenon offers valuable insights into data curriculum design for large-scale model training. Consequently, we believe FastMix can be extended beyond data mixing to serve as a powerful framework for data source attribution. We highly welcome community interest and invite collaboration and further discussion.

\bibliography{iclr2026_conference}
\bibliographystyle{iclr2026_conference}

\appendix

\end{document}